\documentclass[journal]{IEEEtran}
\usepackage{amsmath,amsfonts}
\usepackage{algorithmic}
\usepackage{algorithm}
\usepackage{array}
\usepackage[caption=false,font=normalsize,labelfont=sf,textfont=sf]{subfig}
\usepackage{textcomp}
\usepackage{stfloats}
\usepackage{url}
\usepackage{verbatim}
\usepackage{graphicx}
\usepackage{amsmath}
\usepackage{amssymb}
\usepackage{float}
\usepackage{parskip}
\usepackage{indentfirst}
\usepackage{indentfirst}
\usepackage{cite}
\begin{document}

\title{Multi-scale Fusion Fault Diagnosis Method Based on Two-Dimensionaliztion Sequence in Complex Scenarios}

\author{Weiyang~Jin,~\IEEEmembership{Student Member,~IEEE}
}



\maketitle

\begin{abstract}
Rolling bearings are critical components in rotating machinery, and their faults can cause severe damage. Early detection of abnormalities is crucial to prevent catastrophic accidents. Traditional and intelligent methods have been used to analyze time series data, but in real-life scenarios, sensor data is often noisy and cannot be accurately characterized in the time domain, leading to mode collapse in trained models. Two-dimensionalization methods such as the Gram angle field method (GAF) or interval sampling have been proposed, but they lack mathematical derivation and interpretability. This paper proposes an improved GAF combined with grayscale images for convolution scenarios. The main contributions include illustrating the feasibility of the approach in complex scenarios, widening the data set, and introducing an improved convolutional neural network method with a multi-scale feature fusion diffusion model and deep learning compression techniques for deployment in industrial scenarios. 
\end{abstract}


\section{Introduction}
\IEEEPARstart{R}{olling} bearings play a critical role as components in rotating machinery. Various types of bearing faults can lead to severe damage to the mechanical system. In practical engineering, rolling bearings are often associated with complex faults, where multiple faults coexist simultaneously. Early detection of abnormalities in the bearing can effectively prevent catastrophic accidents during the later stages of fault development. Therefore, it is of utmost importance to monitor and identify potential faults in rolling bearings in their initial stages. Previous research has primarily focused on studying time series data using traditional~\cite{rai2007bearing,yan2014wavelets}or intelligent methods~\cite{eren2019generic,huang2022novel}. However, in complex real-life scenarios, sensor data is often affected by noise and cannot be accurately characterized in the time domain. This can result in mode collapse in trained models~\cite{pan2020two,sun2018gear,yu2007application,wang2010comparative}. To address this issue, some scholars have proposed using two-dimensionalization methods, such as the Gram angle field (GAF) or interval sampling, to create grayscale images~\cite{zhang2019automated,zhang2021fault}. However, these methods are often based on intuition and lack mathematical interpretability and derivation. Furthermore, they only consider consistent accuracy in downstream deep learning tasks, providing little inspiration for further research.

With the continued development of deep learning, the interpretability of convolutional neural networks (CNNs) has become a topic of interest for comparison of time-domain methods and two-dimensional (2D) methods under specific conditions~\cite{sharif2014cnn,yosinski2014transferable}. Recent research has explored the use of heat maps and spectrograms to systematically examine CNNs from the perspective of receptive fields. Additionally, studies have investigated the adaptability of network depth and the upper bound of the generalization error~\cite{zhou2018understanding} of CNNs, providing a theoretical basis for further research in this field. 

Currently, the research and analysis of bearing fault diagnosis mainly concentrate on addressing the problem of coupling of each part of the fault and how to handle it under noisy and real industrial conditions. Two schemes have been proposed to tackle this issue. The first approach involves utilizing the feature fusion method based on U-net~\cite{oktay2018attention} to effectively handle fault problems of varying scales at the fault signal end. The second approach incorporates the attention transfer mechanism~\cite{li2018hierarchical} to regulate the weight of different components in determining the degree of impact of the final fault.

The main contributions of this article are described as
follows:
\begin{enumerate}
\item{Illustrates the feasibility of using a time series two-dimensionalization approach in deep learning fault diagnosis methods, which is shown to be more robust in complex environmental scenarios compared to traditional methods. }
\item{Proposes a method to widen the existing data set scheme to enhance the generalization and migration abilities of the model in different scenarios.}
\item{An improved convolutional neural network method is introduced, incorporating a multi-scale feature fusion to enhance model performance, and utilizing deep learning compression techniques to make the model lightweight and suitable for deployment in industrial scenarios. }
\end{enumerate}

\section{BASIC METHODS}
\subsection{Sequence Two-Dimensionalization}

The Gramian Angular Field (GAF) is a method that can convert time series data into visual representations, utilizing the strengths of machine vision [GAF]. However, the original method produces outputs that are distributed around $0$ and lack sparsity. To address these limitations, this study proposes an improved version of the GAF method that incorporates grayscale images for convolutional scenarios. 

Input vibration sequence can be regarded as $X=x_{1},x_{2},x_{3}...x_{n}$. Simultaneously, in order to utilize deep learning techniques, time series data needs to be transformed into a multi-dimensional space. A bijective mapping is created between the one-dimensional time series and a two-dimensional space to prevent loss of information. 

The Min-Max scaler is used here, so that the sequence is first scaled to $\left [-1,1  \right ] $, then transform to $\left [0,1 \right ] $ through linear mapping.

\begin{equation}\left\{\begin{array}{l}
\phi_{i}=\arccos \left(x_{i}\right) \\
r_{i}=\frac{i}{N}
\end{array}\right.\label{eq:mapping-bijection}\end{equation}where the time series consists of $N$ timestamps $t$ and corresponding $x$. When calculating the radius variable, the interval [0,1] is divided into $N$ equal parts, and then the separation points obtained by using $N$ intervals are associated with the time series.

Subsequently, a novel inner product operation is employed in the two-dimensional space to address the issue of sparsity. Moreover, a significant gap between the constructed two-dimensional matrix and Gaussian noise is observed.
\begin{equation}x \odot  y=x \cdot y-\sqrt{1-x^{2}} \cdot \sqrt{1-y^{2}}\label{eq:dot}\end{equation}
At this point, the matrix can be obtained by taking the outer product expansion of $X$ itself.
\begin{equation}
G=\left(\begin{array}{cccc}
x_{1}\odot x_{1}  & x_{1}\odot x_{2} & \cdots & x_{1}\odot x_{N} \\
x_{2}\odot x_{1}  & x_{2}\odot x_{2} & \cdots & x_{2}\odot x_{N} \\
\vdots & \vdots & \ddots & \vdots \\
x_{N}\odot x_{1} & x_{N}\odot x_{2} & \cdots & x_{N}\odot x_{N}
\end{array}\right)
\end{equation}Next, a type of linear transformation is performed on each element of the matrix $G$ to convert it into an $N\times N$ grayscale image. 
\begin{equation}G'_{ij}=255 \times \frac{G_{ij}+1}{2}\end{equation}

A penalty value is introduced to the dot product matrix to ensure that the resulting map is sparse and does not overlap with Gaussian noise.
\begin{equation}
 f(x,y) =x\odot y-\left \langle x,y \right \rangle=-\sqrt{(1-x^2)(1-y^2)}    \\
\end{equation}
\begin{equation}\left\{\begin{array}{l}
\frac{\partial f(x,y)}{\partial x} =\frac{(1-y^2)x}{\sqrt{(1-x^2)(1-y^2)}} \\ \\
\frac{\partial f(x,y)}{\partial y} =\frac{(1-x^2)x}{\sqrt{(1-x^2)(1-y^2)}} 
\end{array}\right.\end{equation}
Through the partial derivative, it can be seen that the closer $x$ and $y$ are to $0$, the greater the penalty. The main reason is that the points are closer to Gaussian noise.

\subsection{Design of Feature Fusion Network}
The proposed multi-scale feature fusion network has the following distinctive features:
\begin{itemize}
\item{It uses two inputs, grayscale image and time series, to extract features separately, then fuses them, and applies the sigmoid function for binary classification.}
\item{It introduces a feature conversion module to effectively utilize the features of the grayscale image after traversing the ResNet18 network.}
\item{The network integrates both local and global features to better handle complex scenarios.}
\end{itemize}

In order to extract features from time series data, we employ a combination of convolution and multi-layer perceptron, followed by max pooling to obtain global information. Specifically, given an input of size $N\times 1$, we obtain a result of size $64\times 1\times 1$ through this process.

In order to extract features from grayscale images, we take into account the local features of two dimensions. To achieve this, we use one of the input paths as the standard ResNet18 network process, while the other incorporates the result of a feature transformer shown in Fig. 1. 
\begin{figure}[htbp]
\begin{center}
    \centering
   
        \includegraphics[width=2in]{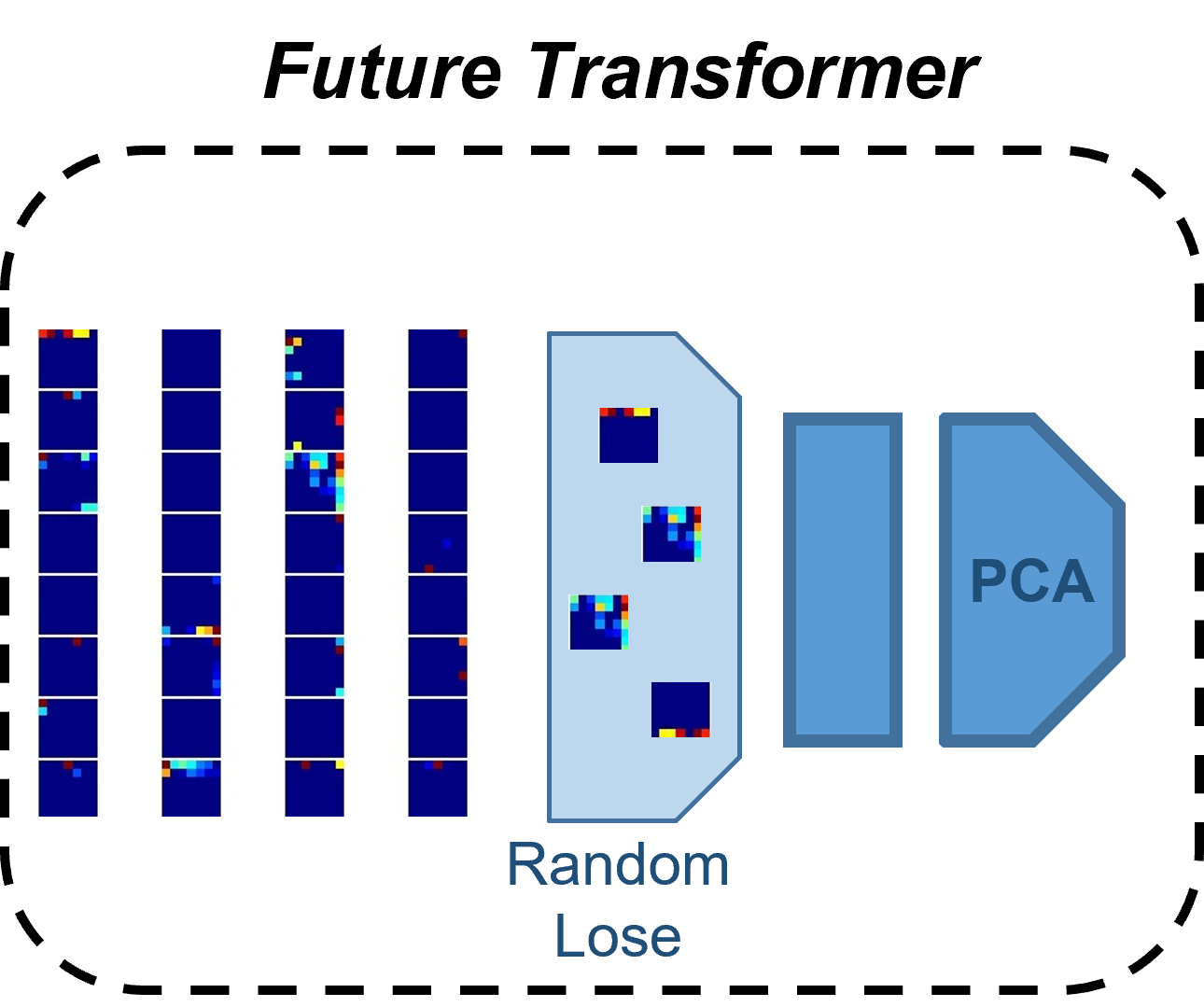}

\end{center}
    \caption{Feature transformer module: Incorporates a random feature loss mechanism to enhance the emphasis on significant features and a module for dimensionality reduction.}

\end{figure}

The processing results for grayscale images can be obtained by utilizing the ResNet network architecture. Given an $N\times 1$ input, the front part of the ResNet network is used to produce an intermediate result $512\times \frac{N}{32}\times \frac{N}{32}$. This result is then further processed through feature extraction using two separate modules, resulting in the final outputs of $512\times 1\times 1$ and $512\times \frac{N}{32}\times 1$. In order to classify the signals, we combine the three types of feature extraction results through a feature layer and then fuse them using a fully connected layer. The resulting features are then passed through a sigmoid function to classify the signals. 

To optimize the proposed network, we take into account both global and local information by using loss functions from two perspectives. The global information is derived from the original sequence, while the local information is extracted from the image. Since the ultimate goal is to classify the signal, we employ the cross-entropy logarithmic loss function for transformation. In this context, $N$ represents the total number of samples in a dataset, $z_{n}$ represents the predicted score for the nth sample being a positive example, $y_{n}$ represents the corresponding true label for the nth sample, and the sigmoid function is denoted as $\delta$. We have
\begin{equation}\begin{array}{l}
\\ \operatorname{loss}(z, y)=\operatorname{mean}\left\{l_{0}, \ldots l_{N-1}\right\} \\
\\
l_{n}=-\left(y_{n} \log \left(\delta\left(z_{n}\right)\right)+\left(1-y_{n}\right) \log \left(1-\delta\left(z_{n}\right)\right)\right)
\end{array}\end{equation}

\begin{figure*}[!t]
\centering
  \includegraphics[width=5in]{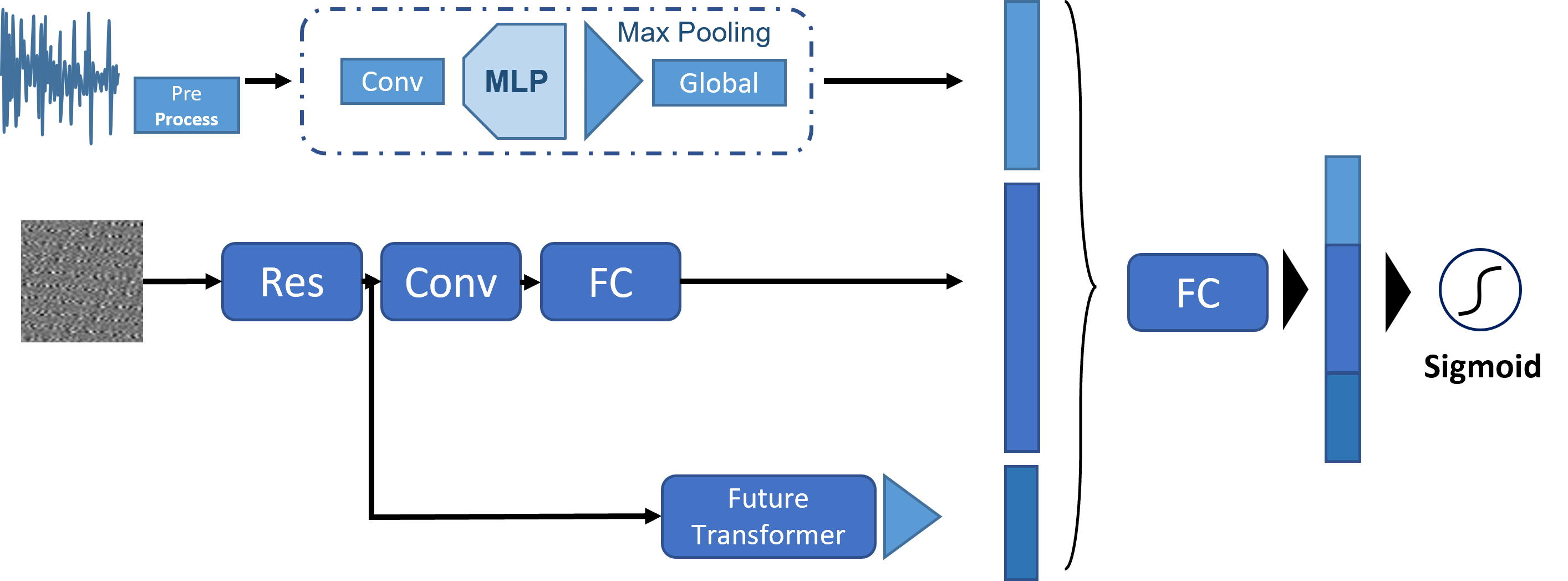}
\caption{Multi-scale feature fusion network, including two types of input and three-dimensional feature extraction methods.}
\label{fig2}
\end{figure*}
\subsection{Model Deployment}
In our model, the main pruning modules are concentrated in the ResNet network. However, pruning ResNet poses some challenges due to its shortcut structure~\cite{molchanov2019importance}. The last convolutional layer in the bottleneck cannot be pruned because it must be consistent with the number of channels in the branch, which means that the number of channels cannot be reduced. To overcome this challenge, we use a Channel Selection function, which filters channels by setting them to $0$ or $1$. This function allows us to block certain channels without actually reducing the number of channels, resulting in pseudo-pruning.

Channel Selection is necessary for pruning ResNet because the structural formula of the product block is BN - ReLU - Conv, and the first single convolution cannot be pruned as there is no BN layer in front of it. If we were to prune the first BN layer of the bottleneck according to the weight, then the channel numbers between the two layers would be incorrect and unable to be connected in series. Therefore, the first BN of the bottleneck can only be pseudo-pruned through Channel Selection. Additionally, there may be some separate convolutions connected between bottlenecks, which cannot be pruned due to the lack of the associated BN layer. Therefore, the first BN layer of the latter bottleneck also needs Channel Selection to ensure that it has the same dimension as the previous Conv layer.

Finally, we consider BN pruning, and the relevant results will be reflected in the experimental module.

\section{EXPANSION OF THE DATASET}
We have used the bearing dataset from Case Western Reserve University as the primary dataset for our research~\cite{smith2015rolling}. To address the limitation of the dataset's capacity and the insufficient number of feature types, we have employed the following approaches to augment the signal:

\begin{itemize}
\item{Time series signal data set expansion based on DCGAN~\cite{radford2015unsupervised}.}

\item{Noise the original dataset to increase the signal-to-noise ratio.}
\end{itemize}

\subsection{DCGAN to Expansion}
\emph{Step 1}: Transform the one-dimensional time series signals of normal data, rolling element fault data sets, outer ring faults, and inner ring faults into grayscale images.

\emph{Step 2}: In order to process the data using DCGAN, convert it into an image format and then input it into the network. It is important to note that the images should be imported in batches during this process to improve efficiency and speed up the training process.

\emph{Step 3}: When initializing the training process of the DCGAN model, the label of the real image is assigned the value of $1$, while the label of the fake image is assigned the value of $0$. However, it is possible to use floating-point values within a certain range for the labels.
\begin{figure}[htbp]
\begin{center}
    \centering
   
        \includegraphics[width=3in]{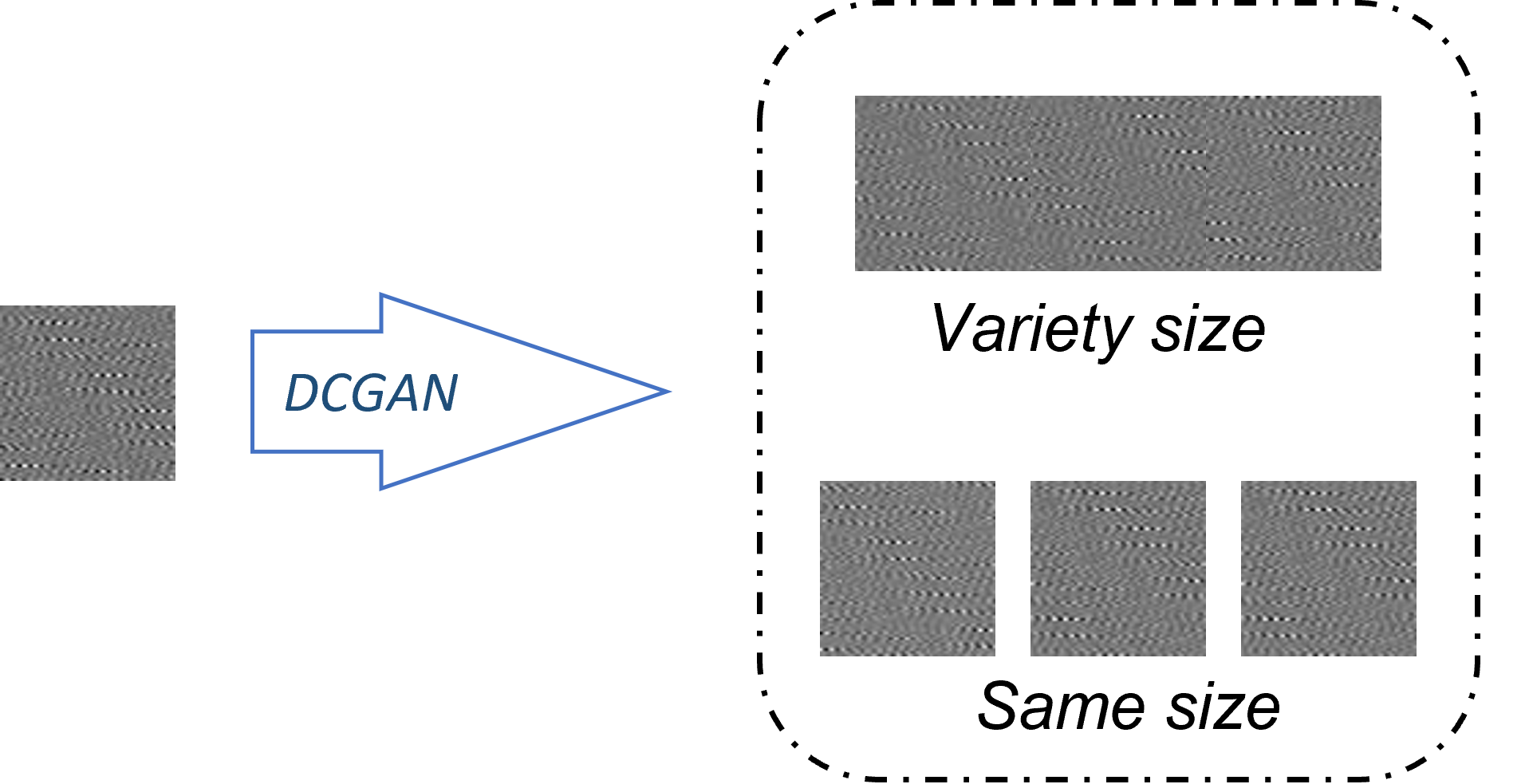}

\end{center}
    \caption{Using DCGAN to generate results: different size specifications, with adaptability.}
    \label{fig.3}
    
\end{figure}
Here, in Fig. 3, we observe that DCGAN is able to control the size of the generated image, which implies that it can be employed in a wider range of data set expansion modes. This, in turn, can enhance the generalization capability of the model and make it applicable to various diagnostic scenarios.

\subsection{Noise Addition Extension}
To improve the robustness of the model, one effective approach is to increase the noise content in the source data. This can be achieved by reducing the signal-to-noise ratio of time series data or adding disturbance points to grayscale images during the construction process.

The noise percentage $\epsilon$ has the following relationship with the signal-to-noise ratio $\mathrm{SNR}$
\begin{equation}
\mathrm{SNR}=20 \times \log _{10}\left(\frac{1}{\varepsilon}\right)
\end{equation}

The percentage of noise in a signal can be measured as either the ratio of noise amplitude to signal amplitude, or the square root of the ratio of noise power to signal power. However, due to high-frequency vibration interference, measuring the amplitude is not always accurate. Therefore, it is more reliable to use the power or effective value of the signal to calculate the noise percentage.
\begin{equation}
\sqrt{\frac{\mathrm{P}_{\mathrm{N}}}{\mathrm{P}_{\mathrm{S}}}}=\sqrt{\frac{\mathrm{RMS}_{\mathrm{N}}^{2}}{\mathrm{RMS}_{\mathrm{S}}^{2}}}=\frac{\mathrm{RMS}_{\mathrm{N}}}{\mathrm{RMS}_{\mathrm{S}}}
\end{equation}
In order to ensure that the noise added to the data follows a zero-mean Gaussian distribution, we can use the effective value of the zero-mean noise, which is equivalent to its standard deviation. However, the percentage value $\epsilon$ of the noise is a relative quantity and therefore cannot be directly used as the variance $\sigma$ of the Gaussian distribution, which is a quantitative class. To determine the variance of the noise distribution, we can use equations (8) and (9)
\begin{equation}
\sigma_{\mathrm{N}}=\varepsilon \cdot \mathrm{RMS}_{\mathrm{S}}
\end{equation}

To noise series $X=x_{1},x_{2},x_{3}...x_{n}$, it can follow these steps: Use (10) to find the variance, and then perform Gaussian sampling. After the noise signal is constructed, it is directly added to the original signal to obtain a new sample.
\begin{figure}[htbp]
\begin{center}
    \centering
   
        \includegraphics[width=2in]{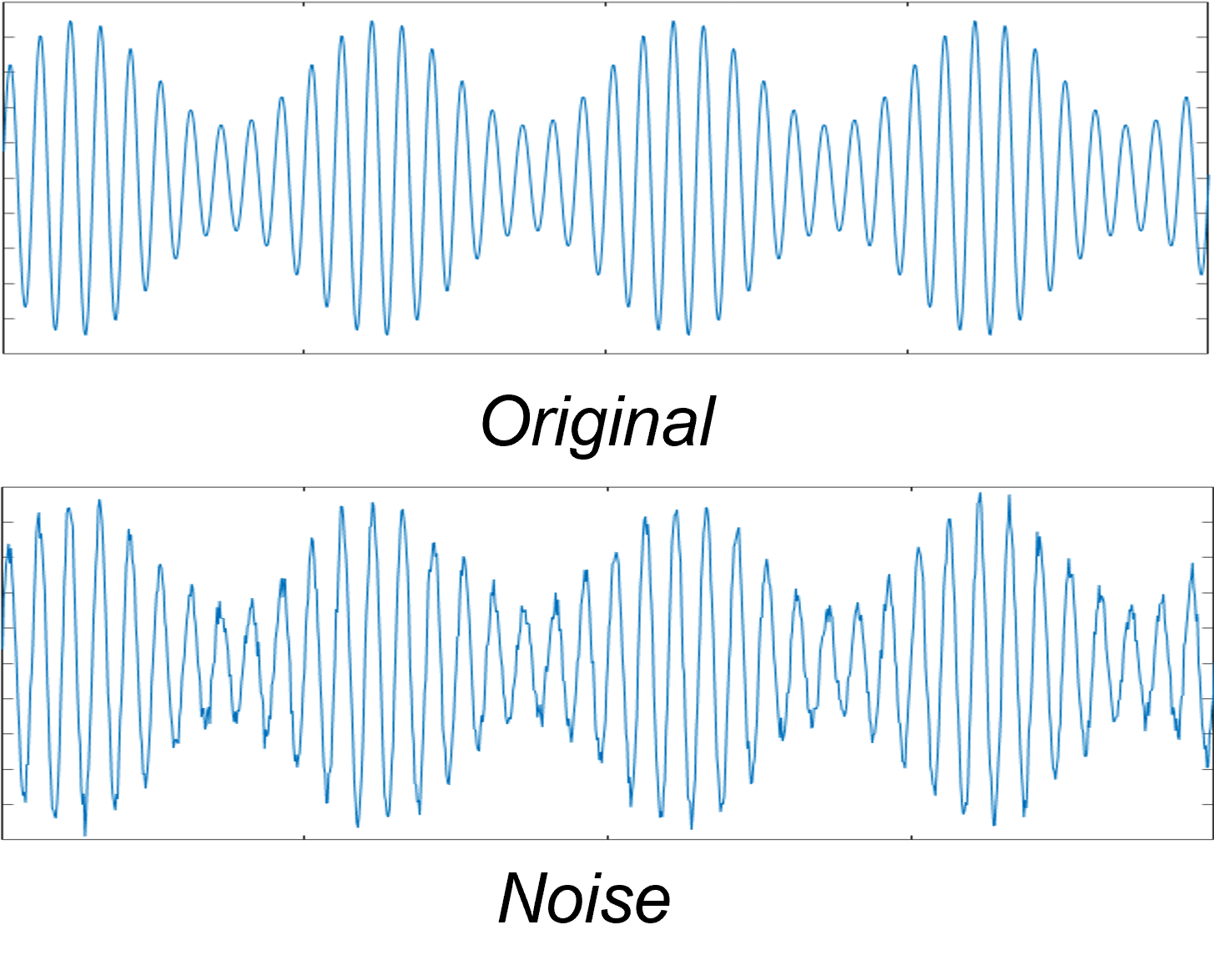}

\end{center}
    \caption{The original signal and the signal after adding noise.}
    \label{fig.4}
    
\end{figure}
For adding noise to the image, we can use the idea of diffusion model~\cite{ho2020denoising,nichol2021improved} and use the construction of Markov process. For the original grayscale image $x_{0}$, gradually add noise. Through this process, there can be recursive results
\begin{equation}\begin{array}{l}
x_{t}=\sqrt{\alpha_{t}} x_{t-1}+\sqrt{1-\alpha_{t}} \epsilon_{t-1}\\ \\
x_{t}=\sqrt{\overline{\alpha_{t}}} x_{0}+\sqrt{1-\overline{\alpha_{t}}} \epsilon
\end{array}\end{equation}$x_{t}$is the signal after adding noise, among them, $\left\{\alpha_{t}\right\}_{t=1}^{T}$ is a pre-set hyperparameter, called Noise schedule, usually a series of small values. $\epsilon_{t-1} \sim N(0,1)$ and $\epsilon \sim N(0,1) $ are Gaussian noise.

The derivation process can be written as \begin{equation}
    \begin{array}{l}
\mathbf{x}_{t}=\sqrt{\alpha_{t}} \mathbf{x}_{t-1}+\sqrt{1-\alpha_{t}} \epsilon_{t-1} \\
=\sqrt{\alpha_{t}}\left(\sqrt{\alpha_{t-1}} \mathbf{x}_{t-2}+\sqrt{1-\alpha_{t-1}} \epsilon_{t-2}\right)+\sqrt{1-\alpha_{t}} \epsilon_{t-1} \\
=\sqrt{\alpha_{t} \alpha_{t-1}} \mathbf{x}_{t-2}+\sqrt{{\sqrt{\alpha_{t}-\alpha_{t} \alpha_{t-1}}}^{2}+{\sqrt{1-\alpha_{t}}}^{2}} \bar{\epsilon}_{t-2} \quad \\ \\\text { where } \bar{\epsilon}_{t-2} \text { merges two Gaussians }(*) . \\\\
=\sqrt{\alpha_{t} \alpha_{t-1}} \mathbf{x}_{t-2}+\sqrt{1-\alpha_{t} \alpha_{t-1}} \bar{\epsilon}_{t-2} \\
=\ldots \\
=\sqrt{\bar{\alpha}_{t}} \mathbf{x}_{0}+\sqrt{1-\bar{\alpha}_{t}} \epsilon \\
\end{array}
\end{equation}
\begin{figure}[htbp]
\begin{center}
    \centering
   
        \includegraphics[width=2in]{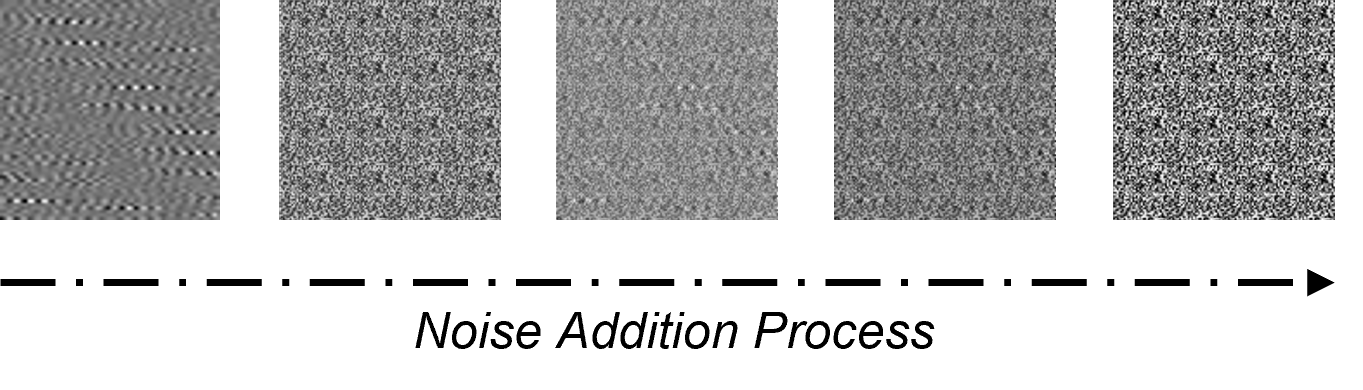}

\end{center}
    \caption{With a gradual noise addition process, it can eventually become white noise, but we only take it to the right case.}
    \label{fig.5}
    
\end{figure}
\section{EXPERIMENT}
This experiment includes the following parts:
\begin{itemize}
\item{Test set evaluation of different models based on the Case Western Reserve University dataset.}

\item{Test accuracy evaluation by adding ambient real noise, including sequence noise and white Gaussian noise.}
\item{The decline of test accuracy and the reflection of the Average precision~\cite{robertson2008new,he2018local} after different pruning degrees.}
\end{itemize}
\subsection{Experimental Design}
The comparison experiment will be compared with the optimal model of the data set based on ResNet-based TCNN, including the comparison at the level of robustness and accuracy.

For noise evaluation on real scenes, our experimental design is as follows:
\begin{figure}[htbp]
\begin{center}
    \centering
   
        \includegraphics[width=3in]{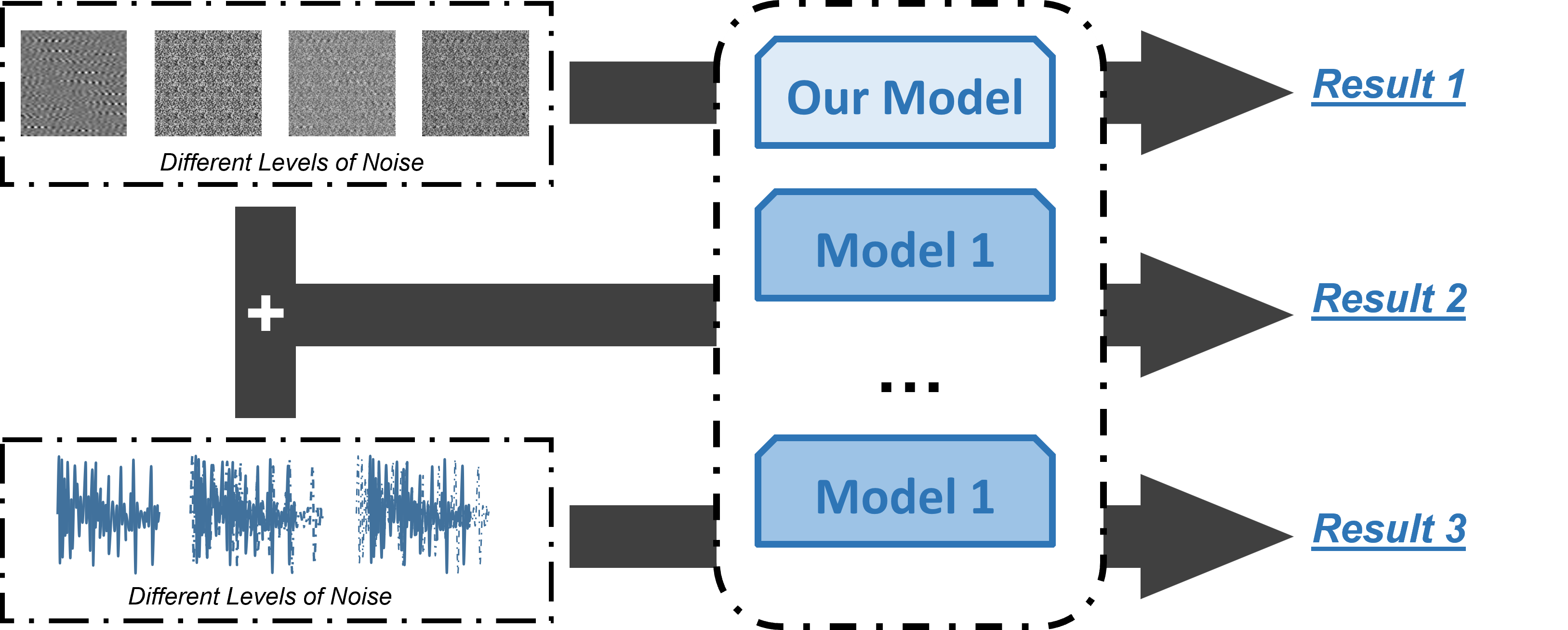}

\end{center}
    \caption{Test results of different models after adding different levels of noise.}

\end{figure}
As shown in Fig.6, we add Gaussian noise from two dimensions for testing, and compare and analyze the results in the two dimensions of accuracy and average precision.
\subsection{Training and Optimization}
$\textbf{Training}$. The training process roughly conforms to the model of the convolutional neural network category. It should be noted that the preprocessing module and dimensionality reduction module in the feature extraction process.
$\textbf{Optimization}$. In order to draw conclusions with higher confidence, we modified the training samples by randomly flipping and adding noise to construct a mini-batch of size 32. The Adam~\cite{zhang2018improved} optimizer is used with an initial learning rate of $10^{-4}$ and the learning rate is multiplied by a factor of 0.8 every 3 epochs.
\subsection{Model Comparison Experiment}
Here, $5\%$, $10\%$, $20\%$, and $50\%$ time series signal noise in four dimensions are used to compare the results, including the traditional method FFT and WDCNN.

In Fig.7 and Table 1, a comparison is presented regarding the robustness and accuracy of different models when faced with real environmental noise. It is observed that for the original data, both the TCNN method and our model have obtained the best results. However, as the noise ratio increases, our model shows good robustness, which is not the case with the TCNN method. Although traditional methods have a relatively good performance in terms of noise robustness, they are not a favorable choice since their accuracy is relatively low compared to the proposed model.

\begin{table}[htbp]
\caption{Accuracy under different noises, in percentage. Bold indicates best in this comparison}
\centering
\begin{tabular}{llllll}
\hline
\textbf{}          & Origin & 5\% & 10\%  & 20\%  & 50\%  \\ \hline
\textbf{Our Model} & \textbf{99.5}  & \textbf{99.5} & \textbf{99.3} & \textbf{98.2} & \textbf{95.5} \\
WDCNN~\cite{zhang2017new}               & 96.1           & 93.0          & 90.7          & 86.6          & 83.2          \\
TCNN(ResNet)~\cite{wen2020transfer}         & \textbf{99.5}  & 99.2          & 98.0           & 91.0          & 88.2          \\
FFT~\cite{rai2007bearing}                 & 89.1           & 89.1          & 88.1          & 82.1          & 71.2          \\
SVM~\cite{shuang2007bearing}                & 89.7           & 88.7          & 87.2          & 85.9          & 82.1          \\ \hline
\end{tabular}
\end{table}
\begin{figure}[htbp]
\begin{center}
    \centering
   
        \includegraphics[width=3in]{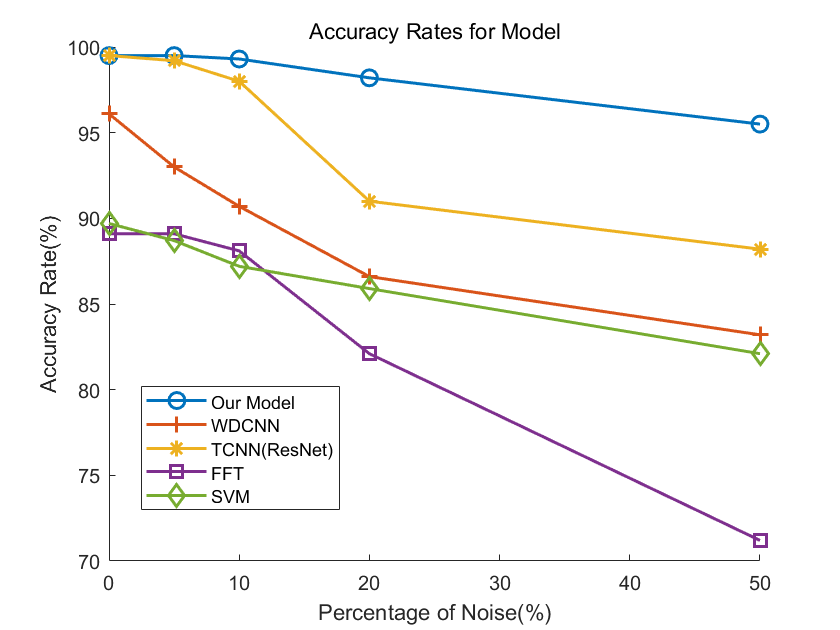}

\end{center}
    \caption{Accuracy under different noises.}

\end{figure}

\subsection{Robustness Experiment}
The robustness experiment aims to evaluate the performance of a model in complex scenarios, where the model may encounter various challenges, such as noise, occlusion, and other disturbances. The evaluation of robustness is usually based on metrics that measure the accuracy and generalization ability of the model in different scenarios. In this study, we use two evaluation metrics, namely accuracy and average precision, to measure the robustness of different models. Accuracy measures the proportion of correctly classified samples in the test set, while average precision measures the quality of the ranked results, taking into account both precision and recall. By comparing the performance of different models in real environmental noise, we can assess their robustness and select the most suitable model for the given task.
\begin{figure}[htbp]
\begin{center}
    \centering
   
        \includegraphics[width=3in]{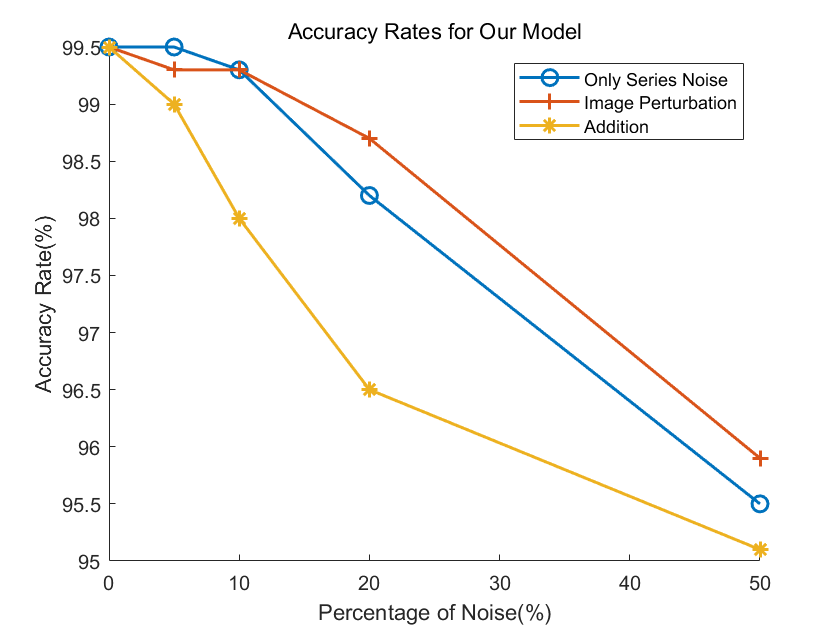}

\end{center}
    \caption{Accuracy under different noises.}

\end{figure}
The Fig.8 illustrates that when the input data is contaminated with noise, the model's performance significantly deteriorates. This degradation can be attributed to the loss of both local and global features due to the noise, which leads to the incorrect extraction of features and subsequent misclassification.

For the calculation of the AP value in this case, we frame the whole with the largest rectangle, and then calculate the percentage of the area under the broken line in the rectangle
\begin{equation}
    AP=\frac{S_{line} }{S_{rectangle}} = \frac{\sum_{i=1}^{N}\frac{(x_{i}-x_{i-1}x)(y_{i}+y_{i-1})}{2} }{x_{max}\times y_{max} } 
\end{equation}

In accordance with equation (13), the average precision (AP) value can be calculated for each noise condition, which is then used as a metric to evaluate the model's robustness performance.
\begin{table}[htbp]
\caption{AP Value for Different Noise.}
\centering
\begin{tabular}{|l|l|}
\hline
          & AP             \\ \hline
Only Series Noise  & 0.9824\\
Image Perturbation & \textbf{0.9854 }         \\
Addition           & 0.9719 \\ \hline
\end{tabular}
\end{table}
The Table 2 demonstrates that our model exhibits strong robustness against image perturbations. This can be attributed to the utilization of a feature extraction approach that incorporates both local and global features. Consequently, even when the image is perturbed by noise, the model can still access a global normal input, enabling it to adjust to such extreme scenarios.

\subsection{Pruning Experiment}
In our pruning scheme for ResNet, we mainly focus on two aspects~\cite{wang2023progressive}. Firstly, a local-based method is used, which involves selecting and removing filters that have the closest local relationship. This means that if a filter is similar enough to its adjacent filters, it can be deleted without impacting the model's performance. Secondly, we aim to retain as much representational power of the pre-trained model as possible while pruning it.

\begin{table}[htbp]
\caption{ACCURACY UNDER Pruning}
 \centering
\begin{tabular}{l|lllll}
\hline
Pruning Rate(\%) & 0    & 10   & 20   & 50   & 90   \\ \hline
Accuracy(\%)     & 99.5 & 99.1 & 98.7 & 98.5 & 95.2 \\ \hline

\end{tabular}
\end{table}
\begin{figure}
\begin{center}
    \centering
   
        \includegraphics[width=3in]{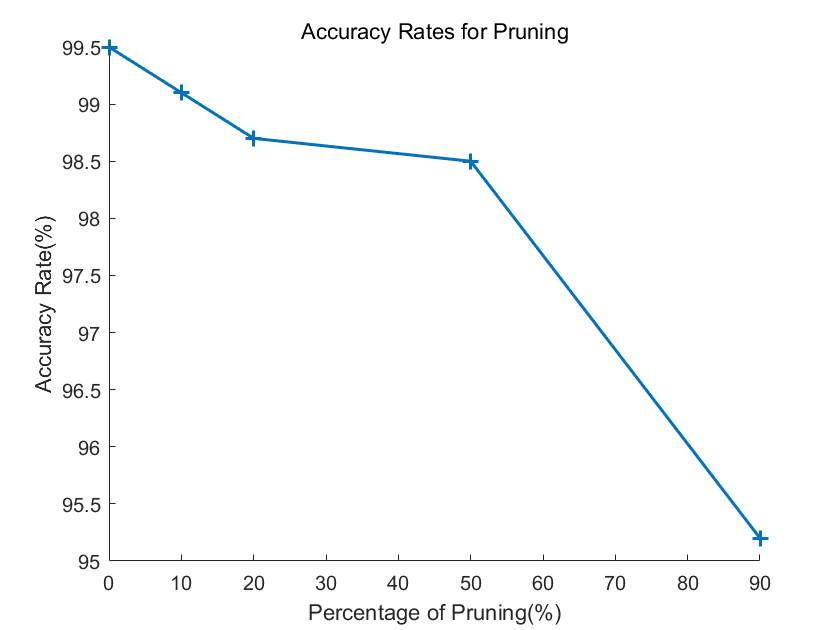}

\end{center}
    \caption{Accuracy under different pruning rate.}
    \label{fig.6}
    
\end{figure}
The experiment showed that even after pruning $90\%$ of the filters, the model still maintained a high level of expressiveness. This suggests that the pruning method can be used to deploy the model on devices with limited memory, making it more practical in real-world situations.

At the same time, for this kind of pruned model, we continue to test with different types of noise to find out whether the pruned model continues to have strong robustness.
\begin{figure}[htbp]
\begin{center}
    \centering
   
        \includegraphics[width=3in]{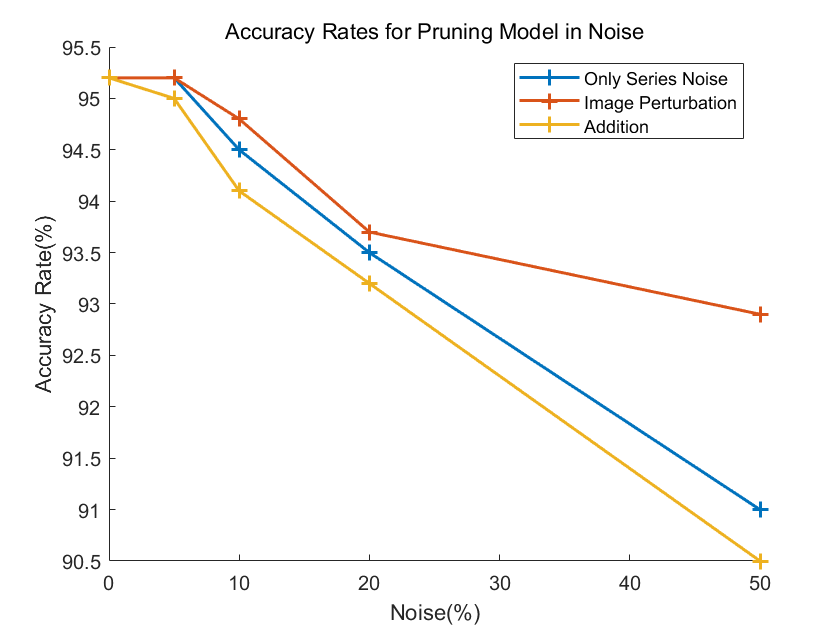}

\end{center}
    \caption{Accuracy under different Noise. Typically, it shows the robustness by accuracy. Also, AP can be used to evaluate it.}

\end{figure}

In the presence of real-world noise, our findings show that the pruned model still exhibits a significant degree of robustness. This may be due to the fact that the pruning is focused on the ResNet skeleton rather than other imported modules, enabling it to maintain effective feature extraction capabilities. The crucial global and local information is still well-extracted by the network, which enables the model to perform well in this aspect.

\subsection{Ablation Study}
In an ablation study, different modules of a system are removed to observe the effect on the overall performance and gain insights into the underlying causal relationships. In the case of our model, we have three main modules: a sequence global information module, two local information feature extraction modules, and a feature fusion module. By removing one of these modules and comparing the results with the original model, we can determine the importance and contribution of each module and assess whether the ResNet backbone model has been significantly affected by the removal of a module.
\begin{table}[htbp]
\caption{Accuracy under different noises}
\centering
\begin{tabular}{lcccc}
\hline
\multicolumn{1}{c}{Model} & Original Accuracy(\%) & 50\% Series Noise(\%)  \\ \hline
Whole                     & 99.5                  & 95.5                                \\
Remove Global Feature     & 99.5                  & 90.5                               \\
Remove Local Feature 1    & 99.5                  & 94.8                      \\
Remove Local Feature 2    & 99.5                  & 94.9                  \\
ResNet Backbone           & 99.5                  & 90.3                       \\ \hline
\end{tabular}
\end{table}

\begin{table}[htbp]
\caption{Accuracy under different noises\label{tab:table1}}
\centering
\begin{tabular}{lcccc}
\hline
\multicolumn{1}{c}{Model}  & Image Perturbation(\%) & 50\% Addition(\%) \\ \hline
Whole                                       & 95.8                   & 95.1              \\
Remove Global Feature                       & 95.6                   & 89.9              \\
Remove Local Feature 1                      & 95.7                   & 94.2              \\
Remove Local Feature 2                      & 95.6                   & 94.3              \\
ResNet Backbone                             & -                      & -                 \\ \hline
\end{tabular}
\end{table}

The ablation study allows for an analysis of the different modules and their contributions to the robustness to noise under various conditions. The global and local information feature extraction modules, although not significantly improving the overall model accuracy, play a role in enhancing the model's robustness to noise. The ResNet backbone remains the primary source of accuracy, while the three modules contribute to the model's robustness under different noise scenarios. Specifically, the sequence global feature extraction module exhibits strong resistance to overall noise, as it maintains the global information despite image disturbance. Meanwhile, the local feature module aids in complementing or assisting the sequence feature information when the global feature is not properly extracted, thus preventing significant accuracy reduction.

\section{Analysis and Optimization}
\subsection{Analysis of Images and Sequence Signals}
Kullback-Leibler Divergence (KL divergence) is a measure of how different two probability distributions are from each other~\cite{goldberger2003efficient}. In this case, we can use KL divergence to estimate the difference between the distribution of the final classification features of the image and the sequence signal during feature extraction. By calculating the KL divergence between these two distributions, we can get an idea of how much information is lost or gained during feature extraction and whether the two inputs are contributing equally to the final classification result.
\begin{equation}\begin{array}{l}
    Series:X\to S_{\theta} (X\mid j_{1}, j_{2},\ldots,j_{n})\to p(X) \\\\
Image:Y\to S_{\theta} (Y\mid k_{1}, k_{2},\ldots,k_{n})\to q(Y)
\end{array}
\end{equation}
For two different distributions, use it to directly bring in
\begin{equation}
    D_{K L}(p \| q)=\sum_{i=1}^{N} p\left(x_{i}\right)\left(\log \frac{p\left(x_{i}\right)}{q\left(y_{i}\right)}\right)
\end{equation}
Using this method, the difference between local information and global information can be well estimated.
\begin{figure}[htbp]
\begin{center}
    \centering
   
        \includegraphics[width=3in]{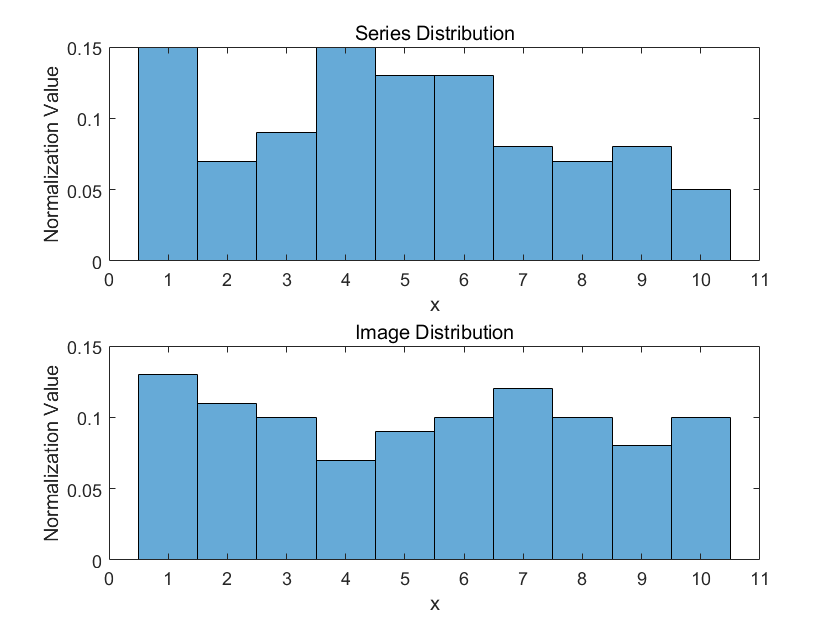}

\end{center}
    \caption{Distribution for image and series. In this case, $D_{K L}(p \| q)=0.2699$}

\end{figure}
\subsection{Noise and Fault Signals}
\begin{figure}[htbp]
\centering
\subfloat[]{\includegraphics[width=2in]{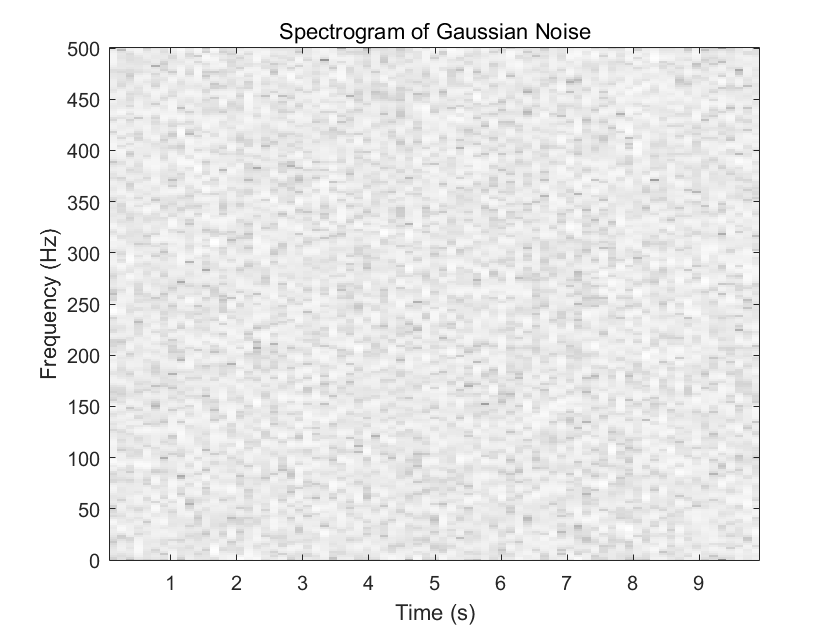}%
\label{fig_first_case}}
\hfil
\subfloat[]{\includegraphics[width=2in]{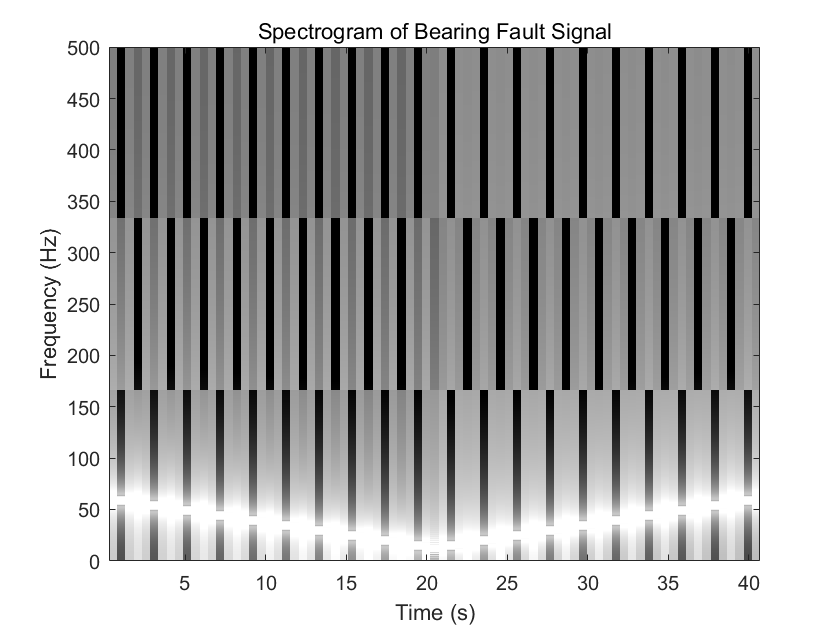}}%
\hfil
\subfloat[]{\includegraphics[width=2in]{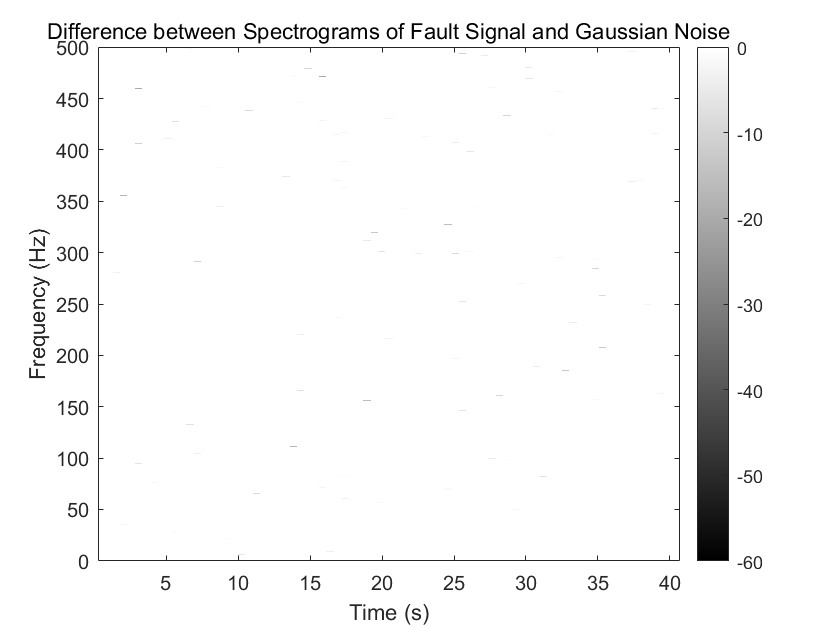}}%
\label{fig_second_case}
\caption{Difference between the Spectrogram of the fault and the spectrum of Gaussian noises. (a) Gaussian noise. (b) Fault signalI. (c) Difference.}
\label{fig_sim}
\end{figure}
The construction of a Gaussian distribution to fit the noise signal is a common practice in signal processing. However, in real-world scenarios, it is crucial to consider the characteristics of the fault signal and how white noise affects it. As such, it is essential to optimize the noise signal model to better reflect the actual noise present in the system. Further research is needed to determine how different types of noise may impact fault detection and how noise models can be improved to better represent real-world scenarios.

The initial step is to extract the differential component of the fault signal, which can be obtained by subtracting the normal signal from the fault signal.

Then, perform Fourier transform on the subtracted result to obtain the corresponding spectrum, and the spectrogram of Gaussian noise can be derived by derivation. Assume that the noise is Gaussian white noise with length $N$ and standard deviation $\sigma$, the discrete Fourier transform of this noise is\begin{equation}
    X[k]=A[k]+j \cdot B[k]=\sum_{n=0}^{N-1}(a[n]+j \cdot b[n]) \cdot e^{-2 \pi jn k/N}
\end{equation}By decomposing the real part and imaginary part, we can get\begin{equation}
    \begin{array}{l}
A[k]=\sum_{n=0}^{N}\left(a[n] \cos \frac{2 \pi n k}{N}+b[n] \sin \frac{2 \pi n k}{N}\right) \\\\
B[k]=\sum_{n=0}^{N}\left(b[n] \cos \frac{2 \pi n k}{N}-a[n] \sin \frac{2 \pi n k}{N}\right)
\end{array}
\end{equation}
After performing the Laplace transform, using the summation formula of independent random variables, it can be directly obtained
\begin{equation}
M_{A[k]}(s)  =\prod_{n=0}^{N-1} M_{r[n]}(s)
 =e^{\frac{s^{2} \sigma^{2} N}{2}}
\end{equation}
Then do the inverse transformation to get the PDF of $A[k]$
\begin{equation}
    f_{B[k]}(x)=\frac{1}{\sqrt{2 \pi N} \sigma} e^{-\frac{x}{2 N \sigma^{2}}}
\end{equation}
The spectrum of the noise can be obtained as Gaussian white noise with a standard deviation of $\sqrt{N}\sigma$.

In the context of the frequency spectrum, distinguishing between a fault signal and Gaussian noise can be challenging, as they can be superimposed and difficult to differentiate using traditional methods. Therefore, traditional methods may not be effective in diagnosing faults directly in this case. Instead, it may be more effective to use traditional methods for preprocessing, such as applying a filter to remove the Gaussian noise before analyzing the fault signal.
\subsection{Analysis of Experimental Results}
In this study, we conducted experiments to compare the robustness of our network under different deployment conditions, with AP serving as an indicator to assess its performance in various situations and models. Through an ablation study, we were able to infer the functions of each module of the network. Our proposed two-dimensional sequence method alone was found to be insufficient in the multi-scale feature fusion method, which aligns with the results of the ablation study. We found that global information feature extraction is more effective in achieving robustness. However, by incorporating local information extracted from grayscale images, we were able to indirectly enhance the capture of global information, resulting in better performance. We conclude that this indirect effect has good performance ability and should be considered in future network designs.

For the pruning experiment, it is important to determine the optimal pruning percentage.
\begin{figure}[htbp]
\begin{center}
    \centering
   
        \includegraphics[width=3in]{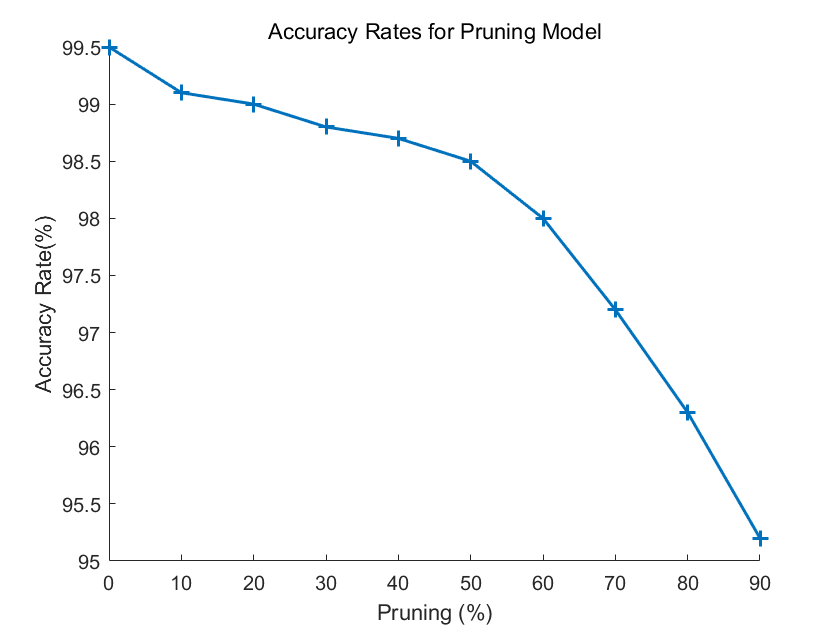}

\end{center}
    \caption{Different percentages of pruning are displayed to find the best pruning result.}
  
\end{figure}

For the values of the two levels: accuracy rate and pruning rate, we expect the higher the better. Therefore, an optimal pruning method can be found by simple Euclidean distance:
\begin{equation}
    I=\sqrt{ACC^{2} +Rate^{2} } 
\end{equation}
For the case where the maximum $I$ value in the above figure is pruned to $90\%$. Since our pruning method is very adaptable, the accuracy rate does not drop too fast with the percentage, making the model tend to the result with a large percentage.
\section{Conclusion}

Aiming at the diagnosis problem of fault diagnosis in complex real scenes, this paper proposes a sequence two-dimensional method to preprocess the input. At the same time, a multi-scale feature fusion network is designed to increase the robustness of fault diagnosis, and then the role of each module is deduced through the ablation study. Aiming at the problem that the model can be deployed on the application side, different percentages of pruning experiments are used to find the possibility of optimizing the combination of pruning rate and learning rate reduction. Then, through the robustness test on the two levels of sequence and image, it is concluded that the model we designed has an amazing performance in terms of robustness in all aspects compared to other models in the past. In terms of data sets, we also propose a new expansion method, using DCGAN for similar supplementation and introducing noise signals to increase the capacity of the data set so as to improve the generalization ability of the fault diagnosis task. Ultimately, we indicate through analysis and optimization, through evaluation, the possibility of future advancement of the overall method.

\newpage

\bibliographystyle{IEEEtran}

\bibliography{references}{}

\end{document}